\documentclass[letterpaper]{article}
\usepackage{aaai2026}
\usepackage{times}
\usepackage{helvet}
\usepackage{courier}
\usepackage[hyphens]{url}
\usepackage{graphicx}
\usepackage{amsmath}
\usepackage{amsfonts}
\usepackage{amssymb}
\usepackage{algorithm}
\usepackage{algorithmic}
\usepackage{natbib}
\usepackage{enumitem}
\title{Predictive Scaling Laws for Efficient GRPO Training of Large Reasoning Models}

\author {
    Datta Nimmaturi\thanks{Joint first authors.}\thanks{Contributed to the early stages of this work, when he was at Nutanix},
    Vaishnavi Bhargava\footnotemark[1],
    Rajat Ghosh\footnotemark[1],
    Johnu George,
    Debojyoti Dutta
}

\affiliations {
    Nutanix Inc. \\
    \texttt{\{vaishnavi.bhargava, rajat.ghosh, johnu.george, debojyoti.dutta\}@nutanix.com},\\
    \texttt{venkatadattasainimmaturi@gmail.com}
}

\begin{document}

\maketitle

\begin{abstract}

Fine-tuning large language models (LLMs) for complex reasoning with reinforcement learning (RL) continues to be prohibitively expensive. Through a phenomenological investigation of GRPO post-training dynamics, we identify a scaling law characterized by exponential reward saturation. The emergence of this early plateau motivates an important question: can GRPO be equipped with principled early stopping criteria to significantly reduce post-training compute while preserving downstream performance?

Across four open-source models—Llama 3B/8B and Qwen 3B/7B—we perform a systematic empirical study of GRPO fine-tuning and derive scaling laws that accurately predict reward trajectories during training. Our analysis shows that GRPO reward curves are well-approximated by an exponential saturation with three phases that are consistent across all models: (i) slow initial progress, (ii) rapid improvement, and (iii) saturation. We further show that a simple parametric scaling law, conditioned on model size, initial performance, and normalized training progress, reliably predicts the onset of plateauing performance.

A key practical finding is that training beyond roughly 80\% of a single epoch yields negligible reward gains while consuming a substantial fraction of total computation. Using our scaling law, practitioners can forecast these phase transitions early and select data-driven stopping points, substantially reducing GRPO compute without sacrificing final performance. Our results suggest that such predictive scaling laws are a promising tool for managing GRPO finetuning costs.

\end{abstract}

\section{Introduction}

Although frontier LLMs such as GPT-5, Claude 3.7, Gemini Ultra 2, DeepSeek-R1, Grok 3, and Kimi K2 exhibit exceptional reasoning in mathematics, coding, and other complex domains, replicating these capabilities internally is difficult for most organizations because the required post-training compute is prohibitively expensive. Post-training methods teach LLMs to perform reasoning-intensive tasks using curated demonstration datasets through either supervised fine-tuning (SFT) or reinforcement fine-tuning (RFT). SFT is primarily a form of behavior cloning, where the model learns to imitate the demonstration distribution \cite{ouyang2022instructgpt, touvron2023llama}. In contrast, RFT—such as Reinforcement Learning from Human Feedback (RLHF) and more recent methods like GRPO—introduces explicit reward signals that incentivize the model to explore and optimize beyond the demonstrations \cite{ziegler2019rlhf, bai2022constitutional, yuan2024grpo}.

Reinforcement fine-tuning (RFT), including techniques such as Proximal Policy Optimization \cite{schulman2017ppo} and Group Relative Policy Optimization \cite{shao2024deepseekmath} has emerged as the most effective method for eliciting complex reasoning capabilities from large language models, enabling them to surpass the limitations of demonstration-based supervised fine-tuning (Ouyang et al. 2022; Ziegler et al. 2019). Recent large-scale systems such as DeepSeek-Math (Shao et al. 2024) and DeepSeek-R1 (Guo et al. 2025) further illustrate that high-performance reasoning models depend critically on costly reinforcement-based post-training such as GRPO. First, RFT requires on-policy sampling, meaning the model must generate fresh rollouts at each training iteration to compute policy gradients. Unlike SFT, where updates are performed over a static dataset, RFT repeatedly invokes the forward pass of the evolving model, often generating multiple candidate responses per prompt to compute comparative or group-based rewards (Bai et al. 2022; Yuan et al. 2024). This repeated sampling can dominate total FLOPs, particularly for long-context reasoning tasks as demonstrated in DeepSeek-Math and DeepSeek-R1 (Shao et al. 2024; Guo et al. 2025). Second, the reward computation pipeline introduces substantial cost. Human preference models, rule-based evaluators, or verifier models must be run on each generated sample to produce reward signals. In multi-sample algorithms such as GRPO, reward-model inference alone can account for 20–40\% of the overall compute budget (Yuan et al. 2024). DeepSeek-Math and DeepSeek-R1 both report that verifier-based reward pipelines are a dominant source of post-training compute. Third, RFT typically requires larger batch sizes and more update steps to stabilize training, especially for reasoning tasks with high reward variance. Policy gradient methods are sensitive to credit assignment noise, which leads practitioners to rely on multiple rollouts per prompt, KL-regularization, and adaptive clipping—all of which increase the effective compute per update (Ziegler et al. 2019; Bai et al. 2022). Finally, RFT amplifies the memory and engineering cost of post-training. Large-scale deployments such as DeepSeek-R1 use pipeline-parallel and expert-parallel execution to jointly run sampling, reward evaluation, and policy updates (Guo et al. 2025). These pipelines are considerably more complex than standard SFT setups (Touvron et al. 2023) and require tight synchronization across GPU clusters. Collectively, these factors make RFT $3–10\times$ more expensive than equivalently sized SFT runs, rendering frontier-level reasoning performance prohibitively costly for most organizations. This motivates the development of scaling laws and early-stopping mechanisms that can reduce RFT computation without degrading final performance—an objective our work directly addresses.

 Early stopping is a well-established regularization technique in deep learning \cite{prechelt1998earlystopping} because it prevents the model from overfitting by halting training before the network begins to memorize noise in the training data. We hypothesize similar possibility exists in GRPO. Although loss curves in GRPO are often interpreted as indicators of optimization progress, it is well-known that performance metrics such as evaluation loss can plateau long before true convergence occurs. In many machine learning systems, the validation reward or accuracy appears to flatten early, even though the underlying optimization dynamics continue to evolve \cite{bottou2012stochastic, goodfellow2016deeplearning}. This discrepancy arises because convergence, in the optimization sense, requires tight stability of parameter updates, whereas plateauing merely reflects diminishing marginal gains in observable metrics. For large language models and reinforcement fine-tuning in particular, the distinction is even sharper: reward curves may reach an early plateau while the policy continues to undergo meaningful parameter shifts driven by KL penalties, variance-reduction effects, or delayed credit assignment. This gap between metric plateau and optimization convergence motivates the need for predictive scaling laws that can detect when further training becomes inefficient.This paper provides a phenomenological investigation into the early plateauing dynamics of GRPO evaluation rewards, highlighting the mismatch between apparent reward stagnation and true optimization convergence.

\section{Preliminary}

Group Relative Policy Optimization (GRPO) \cite{shao2024deepseekmath} is a recent reinforcement learning algorithm designed to improve the efficiency of LLM fine-tuning for reasoning tasks. Traditional Proximal Policy Optimization (PPO) \cite{schulman2017proximalpolicyoptimizationalgorithms} is an actor-critic method that requires training a separate value (critic) model to estimate baselines, introducing significant computational and memory overhead. GRPO eliminates the need for this critic model by estimating baselines from \textit{group-relative advantages}. Specifically, for each input query, GRPO samples a group of outputs and computes the baseline as the average reward within this group, then calculates advantages relative to this baseline. This reduces resource requirements while maintaining stable and effective policy optimization. For each query $q$, GRPO samples a group $G = \{o_1, o_2, \ldots, o_G\}$ of outputs from the old policy $\pi_{\theta_{old}}$ and optimizes the policy $\pi_\theta$ by maximizing:

\begin{multline}
\mathbb{E}_{q \sim P(Q), \{o_i\}_{i=1}^{G} \sim \pi_{\theta_{old}}(O|q)} \Bigg[ \frac{1}{G} \sum_{i=1}^{G} \Big( \min \Big( \frac{\pi_\theta(o_i|q)}{\pi_{\theta_{old}}(o_i|q)} A_i, \\
\text{clipped} \Big( \frac{\pi_\theta(o_i|q)}{\pi_{\theta_{old}}(o_i|q)}, 1-\epsilon, 1+\epsilon \Big) A_i \Big) \\
- \beta D_{KL}(\pi_\theta || \pi_{ref}) \Big) \Bigg].
\end{multline}

Here $A_i$ denotes the advantage computed from a group of rewards $\{r_1, r_2, \ldots, r_G\}$:

\begin{equation}
A_i = \frac{r_i - \text{mean}(\{r_1, r_2, \ldots, r_G\})}{\text{std}(\{r_1, r_2, \ldots, r_G\})},
\end{equation}

and

\begin{equation}
D_{KL}(\pi_\theta || \pi_{ref}) = \frac{\pi_{ref}(o_i|q)}{\pi_\theta(o_i|q)} - \log \frac{\pi_{ref}(o_i|q)}{\pi_\theta(o_i|q)} - 1.
\end{equation}

Note that $\epsilon$ and $\beta$ are parameters controlling the clipping range and KL penalty, respectively.

% \vspace{0.3em}
Furthermore, GRPO supports both \textbf{outcome supervision}, where only final outputs are rewarded, and \textbf{process supervision}, where intermediate reasoning steps receive rewards, providing denser and more informative training signals for reasoning-focused tasks. Empirical results show that GRPO enhances performance compared to PPO while using fewer computational resources, making it especially practical for large reasoning model training.

% \vspace{0.3em}
% In the broader spectrum of RL techniques for LLMs, GRPO complements:

% \begin{itemize}[leftmargin=*]
% \item \textbf{Rejection Sampling Fine-Tuning (RFT)} \cite{yuan2023scalingrelationshiplearningmathematical} : Fine-tuning on outputs filtered by correctness, using offline sampled data.
% \item \textbf{Direct Preference Optimization (DPO)} \cite{rafailov2024directpreferenceoptimizationlanguage} : Optimizing policies based on pairwise preferences without explicit reward models.
% \item \textbf{Online RFT} \cite{agnihotri2025efficientonlinerftplugandplay} : Similar to RFT but with online sampling from the current policy for dynamic adaptation.
% \item \textbf{Proximal Policy Optimization (PPO)} \cite{schulman2017proximalpolicyoptimizationalgorithms} : An actor-critic RL approach optimizing expected rewards with a value baseline and KL constraints for stability.
% \end{itemize}

% \vspace{0.3em}
% These methods differ in data source (offline vs. online sampling), reward specification (rule-based, preference-based, or model-based), and optimization approach. GRPO stands out by combining the sample efficiency of online RL with the computational simplicity of baseline-free algorithms, enabling scalable and cost-effective training of reasoning-capable language models.

% \vspace{0.5em}
% \subsection{Parameter-Efficient Fine-Tuning}

% \vspace{0.3em}
% \begin{itemize}[leftmargin=*]
Low-Rank Adaptation (LoRA) \cite{hu2021loralowrankadaptationlarge} is a parameter-efficient fine-tuning method that inserts small trainable low-rank matrices into pre-trained model weights, enabling task adaptation without updating all parameters and significantly reducing memory requirements. Recent work, such as Tina et al. \cite{wang2025tina}, explores combining GRPO with LoRA for efficient reasoning model training. In this paper, we extend this idea by employing \textbf{LoRA with GRPO on quantized models}, enabling scalable reinforcement learning-based fine-tuning of large reasoning models within practical compute budgets.

\section{Methodology}

To ensure reproducibility and accessibility, we designed our experiments using minimal computational resources while maintaining scientific rigor. Our approach combines LoRA (Low-Rank Adaptation) with GRPO to fine-tune open-source models on publicly available datasets.

\vspace{0.3em}
\subsubsection{Model Selection}

We selected four representative models spanning different sizes and architectural families:

\begin{itemize}[leftmargin=*]
\item \textbf{Llama family} Quantized Llama-3B [unsloth/llama-3.2-3b-instruct-unsloth-bnb-4bit] and  Llama-8B [unsloth/meta-llama-3.1-8b-instruct-unsloth-bnb-4bit]
\item \textbf{Qwen family}: Quantized Qwen-3B [unsloth/qwen2.5-3b-instruct-unsloth-bnb-4bit] and Qwen-7B [unsloth/qwen2.5-7b-instruct-unsloth-bnb-4bit] parameters
\end{itemize}

This selection enables systematic analysis of both architectural differences and scaling effects while maintaining computational feasibility.

\vspace{0.3em}
\subsubsection{Parameter-Efficient Fine-Tuning}

We employ LoRA for parameter-efficient fine-tuning of quantized models, which provides several advantages:

\begin{itemize}[leftmargin=*]
\item \textbf{Reduced memory requirements}: Quantization significantly decreases memory consumption

\item \textbf{Modularity}: Training only a small portion of parameters enables easy toggling of reasoning behavior

\item \textbf{Computational efficiency}: Substantial reduction in training time and resource requirements
\end{itemize}

\vspace{0.3em}
\subsubsection{Training Algorithm}

Our training process utilizes GRPO with the following key characteristics:

\begin{itemize}[leftmargin=*]
\item \textbf{Baseline estimation}: Group-based baseline calculation eliminates the need for a separate critic model

\item \textbf{Reward optimization}: Direct optimization of reasoning task rewards

\item \textbf{Policy constraints}: Proximal policy optimization principles maintain training stability
\end{itemize}

\vspace{0.3em}
\subsubsection{Dataset}

We utilize the OpenThoughts dataset \cite{guha2025openthoughtsdatarecipesreasoning} called Math-lightEval \cite{hendrycksmath2021}, which provides high-quality reasoning examples across diverse problem domains. This dataset choice ensures:

\begin{itemize}[leftmargin=*]
\item \textbf{Reproducibility}: Publicly available and well-documented

\item \textbf{Quality}: Carefully curated reasoning traces

\item \textbf{Diversity}: Broad coverage of reasoning task types
\end{itemize}

\vspace{0.3em}
\subsubsection{Training Setup}

We run all our trainings on single Nvidia A100 with 40GB VRAM. We use unsloth \cite{unsloth} for hardware efficient training and open-r1 \cite{openr1} as reference code for our experiments. A very big problem in performing Reinforcement Learning for LLMs is that one needs to run auto regressive inference to generate samples to then later evaluate. This becomes a bottleneck for any RL training on LLM. To help with that, vLLM \cite{kwon2023efficient} is used as the preferred inferencing engine. This creates an additional memory requirement to load and store the model's weights, KVCache and Activations thus putting additional pressure on GPU RAM. In our observations, more than 90\% of the time is spent generating sample completions. Both the memory requirement and training time go up as the sequence length and number of completions go up. So to make sure that we stay within the constraints, we choose 8192 as the maximum sequence length with 8 completions per prompt.

\vspace{0.3em}
\subsection{Evaluation Methodology}

Our evaluation framework focuses on reward progression throughout training, enabling detailed analysis of learning dynamics. We track model performance at regular intervals and identify key transition points in the learning process.

\subsection*{Theoritical Reward Scaling Law}

We observe that the progression of reward during training follows a consistent power-law behavior with respect to training steps. The normalized reward difference can be modeled as:

\begin{equation}
\frac{R_{\max} - R_t}{R_{\max} - R_{\min}} = N_t^{-\alpha}
\label{eq:theory_scaling}
\end{equation}

where:
\begin{itemize}
  \item $R_t$ denotes the reward at step $t$,
  \item $R_{\max}$ and $R_{\min}$ are the maximum and minimum observed rewards during training,
  \item $N_t$ is the training step
  \item $\alpha \approx 0.3$ is an empirically determined exponent.
\end{itemize}

This relation implies a decelerating improvement in reward as training progresses, reflecting a regime of diminishing returns, as mentioned in \cite{schulman2017proximalpolicyoptimizationalgorithms} too. The power-law structure offers a compact means of modeling learning dynamics and may aid in predicting saturation points or optimizing training schedules.

\section{Results and Analysis}

% \begin{figure}
%     \centering
%     \includegraphics[width=1\linewidth]{all.png}
%     \caption{Reward trajectories across Qwen and Llama models}
%     \label{fig:overall}
% \end{figure}

% \vspace{0.5em}

% \subsection{Universal Learning Patterns}

% ---------- NEW 2×2 FIGURE ----------
\begin{figure*}[t]
    \centering

    % Row 1
    \begin{minipage}{0.48\linewidth}
        \centering
        \includegraphics[width=\linewidth]{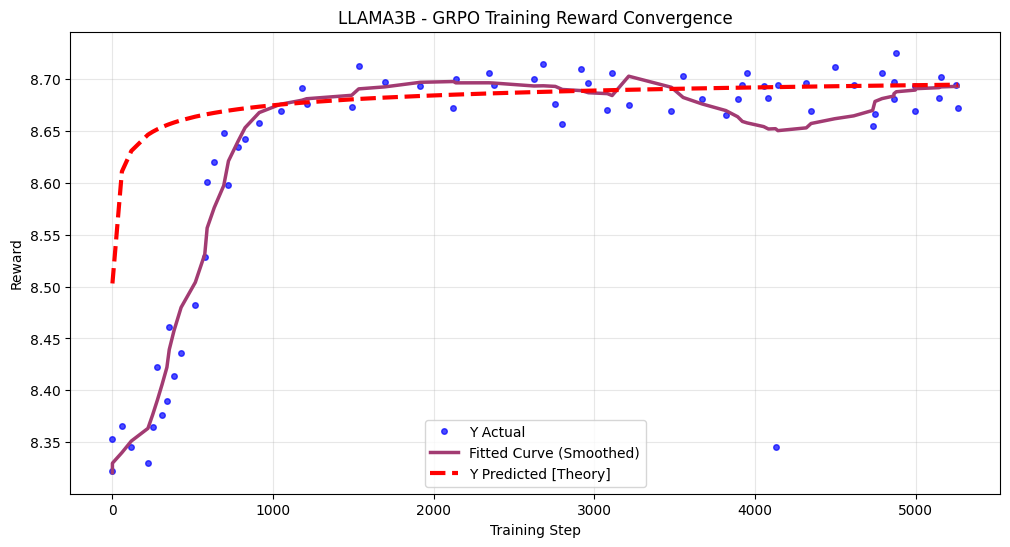}
        \\[-0.4em]
        \small (a) GRPO training trajectory across quantized Llama 3.2 3B model.
    \end{minipage}
    \hfill
    \begin{minipage}{0.48\linewidth}
        \centering
        \includegraphics[width=\linewidth]{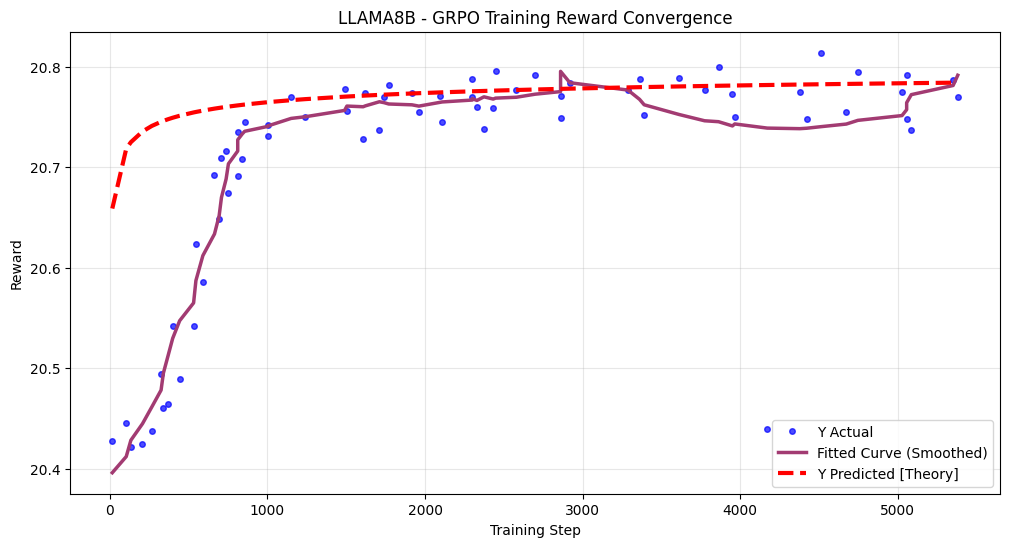}
        \\[-0.4em]
        \small (b) GRPO training trajectory across quantized Llama 3.1 8B model.
    \end{minipage}

    \vspace{0.8em}

    % Row 2
    \begin{minipage}{0.48\linewidth}
        \centering
        \includegraphics[width=\linewidth]{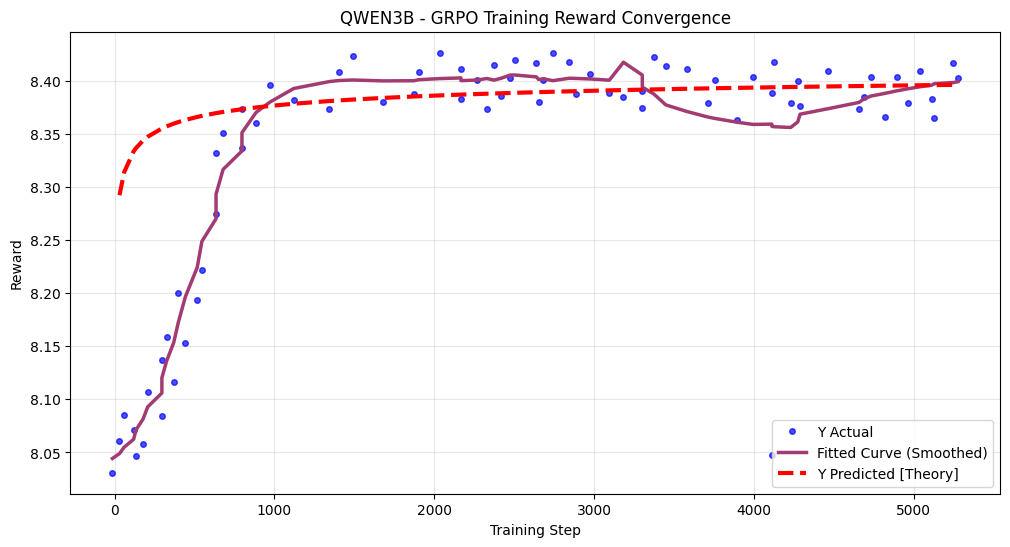}
        \\[-0.4em]
        \small (c) GRPO training trajectory across quantized Qwen2.5 3B model.
    \end{minipage}
    \hfill
    \begin{minipage}{0.48\linewidth}
        \centering
        \includegraphics[width=\linewidth]{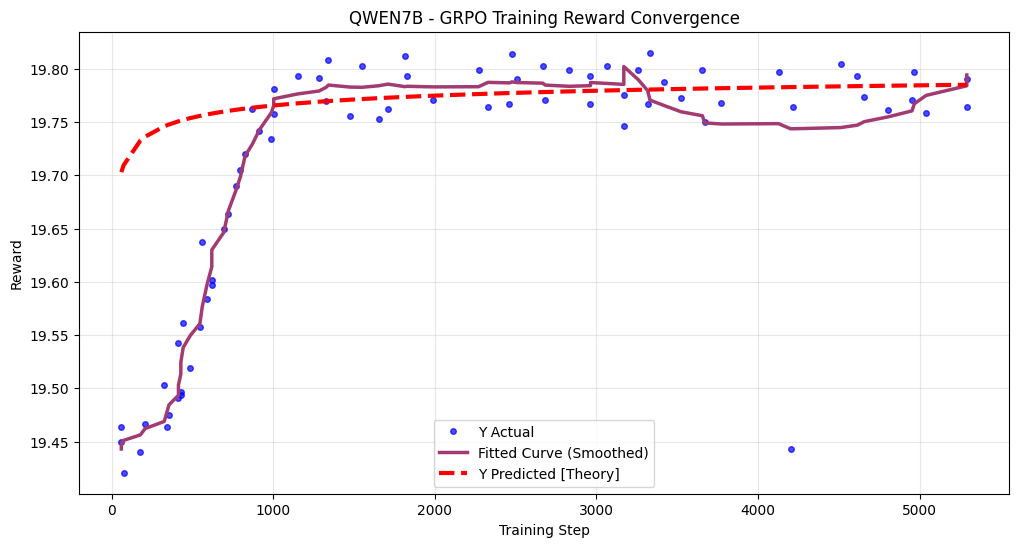}
        \\[-0.4em]
        \small (d) GRPO training trajectory across quantized Qwen2.5 7B model.
    \end{minipage}

    \caption{GRPO training reward convergence across all model configurations. All four models exhibit consistent sigmoid-shaped learning curves with similar phase transitions despite differing parameter counts and architectures.}
    \label{fig:grpo_2x2}
\end{figure*}
% ---------- END 2×2 FIGURE ----------

Our experimental results, as shown in Figure \ref{fig:grpo_2x2}, reveal remarkably consistent learning patterns across all tested models, regardless of size or architectural family. Every model exhibits a distinct sigmoid-shaped (S-shaped) learning curve characterized by three well-defined phases:

\begin{enumerate}[leftmargin=*]
\item \textbf{Initial Phase} (0.0 to 0.1 normalized training steps): A period of slow initial progress where reward increases minimally, suggesting the model is learning reward signal interpretation.

\item \textbf{Rapid Growth Phase} (0.1 to 0.2 normalized training steps): A steep and rapid increase in reward, representing the most efficient phase of learning. The inflection point consistently occurs around the 0.2 mark across all models.

\item \textbf{Plateau Phase} (0.2 onwards): The rate of improvement slows dramatically, and model performance converges to a stable maximum reward level, indicating diminishing returns from continued training.
\end{enumerate}

Hence, we can conclude nearly 80\% of training contribute marginally to the cumulative reward. To quantify the observed learning patterns, we fitted a comprehensive scaling law equation across all models. The empirical scaling law takes the form:

\[
R(t) = \alpha \cdot rinit + \beta \cdot s + \frac{\gamma}{1 + \exp(-\delta \cdot (t - t_0))}
\]

where:
\begin{itemize}[leftmargin=*]
\item $R(t)$ is the reward at normalized training step $t$
\item $r_{init}$ is the initial reward specific to each model configuration
\item $s$ is the model size parameter (in Billions)
\item $\alpha$ is the initial scale factor
\item $\beta$ is the model scale factor
\item $\gamma$ is the sigmoid amplitude factor
\item $\delta$ is the sigmoid steepness factor
\item $t_0$ is the inflection point
\end{itemize}

The fitted global parameters across all datasets are:

\begin{itemize}[leftmargin=*]
\item $\alpha = 1.009$ (initial scale)
\item $\beta = 2.500$ (model scale factor)
\item $\gamma = 0.401$ (sigmoid factor)
\item $\delta = 39.875$ (steepness factor)
\item $t_0 = 0.1$ (inflection point)
\end{itemize}

Across all models, the theoretical curves (Eq.~\ref{eq:theory_scaling}, red dotted lines), empirical fits (dark red solid lines), and observed data points (blue dots) align closely, especially in the plateau regime. This consistency suggests that effective early stopping strategies could be derived for GRPO.

\section{Discussion}

\subsection{Model Size Impact on Performance}

Our results demonstrate a strong correlation between model size and final achievable performance:

\begin{itemize}[leftmargin=*]
\item \textbf{Llama family}: Llama-8B plateaus at approximately 20.8 reward, while Llama-3B reaches only 8.7 reward

\item \textbf{Qwen family}: Qwen-7B achieves approximately 19.8 reward, compared to Qwen-3B's 8.4 reward
\end{itemize}

This 2.4× performance difference between larger and smaller models within each family suggests that parameter count is a dominant factor in determining reasoning capability.

\vspace{0.5em}
\subsection{Architectural Consistency}

Remarkably, models of similar size across different architectural families achieve comparable performance levels:

\begin{itemize}[leftmargin=*]
\item \textbf{Larger models}: Llama-8B (20.8) and Qwen-7B (19.8) show similar final performance

\item \textbf{Smaller models}: Llama-3B (8.7) and Qwen-3B (8.4) converge to comparable reward ranges
\end{itemize}

This observation suggests that model scale is more influential than architectural differences in determining reasoning performance under GRPO training.

\vspace{0.5em}
\subsection{Training Timeline Consistency}

Despite significant differences in final reward achievement, all models follow remarkably similar training timelines:

\begin{itemize}[leftmargin=*]
\item \textbf{Rapid learning initiation}: All models begin steep improvement at approximately 0.1 normalized steps

\item \textbf{Plateau onset}: Performance stabilization occurs around 0.2 normalized steps across all models

\item \textbf{Phase duration}: The rapid growth phase spans approximately 0.2 normalized steps for all models
\end{itemize}

This temporal consistency implies that while the magnitude of learning is determined by model size, the relative speed and phases of learning are universal across the tested architectures.

\subsection{Computational Efficiency Gains}

\vspace{0.5em}
Our findings enable significant computational savings through informed training decisions:

\begin{itemize}[leftmargin=*]
\item \textbf{Early stopping}: Training can be terminated at the 0.2 or 0.3 normalized step mark without meaningful performance loss

\item \textbf{Resource allocation}: Computational resources can be redirected from over-training to exploring different model configurations

\item \textbf{Cost reduction}: Substantial reduction in training costs while maintaining model quality
\end{itemize}

\vspace{0.3em}
\subsection{Model Selection Guidance}

The derived scaling laws provide practical guidance for model selection:

\begin{itemize}[leftmargin=*]
\item \textbf{Performance prediction}: Expected performance can be estimated before training completion

\item \textbf{Resource planning}: Computational requirements can be accurately forecasted

\item \textbf{Architecture selection}: Similar-sized models from different families can be expected to achieve comparable performance
\end{itemize}

\section{Conclusion, Limitation,  and Future Work}

This work provides a detailed empirical and phenomenological examination of GRPO training dynamics for reasoning-oriented language models, uncovering robust and universal patterns that enable more efficient post-training strategies. Across a wide range of architectures and model sizes, we observe consistent saturation-shaped learning curves and develop predictive scaling laws that closely track evaluation reward trajectories. These findings demonstrate that a substantial portion of GRPO’s computational cost occurs well past the point of meaningful reward improvement, indicating that principled early stopping is both feasible and highly effective. By enabling data-driven decisions about training duration and resource allocation, our framework offers a practical pathway toward reducing reinforcement fine-tuning costs without sacrificing model quality—helping to make advanced reasoning capabilities more broadly accessible. The convergence of learning behaviors across models further suggests that the predictive mechanisms identified here may generalize beyond GRPO to other reinforcement learning–based reasoning enhancement methods. While our analysis is grounded in experiments on the OpenThoughts dataset \cite{hendrycksmath2021}, focuses on specific model families and sizes, and operates within a particular set of training configurations, the overarching patterns highlight promising opportunities for future work. These include scaling the analysis to larger architectures, validating universality across diverse reasoning tasks, exploring integration with parameter-efficient fine-tuning techniques, and developing adaptive training schedules that leverage predictive scaling laws. Together, these directions chart a path toward more efficient, accessible, and principled methods for building reasoning-capable LLMs.

\appendix
\section*{Appendix C: Model Response Comparison}

To evaluate the impact of GRPO fine-tuning, we present a representative example from the MathEval dataset comparing responses generated by the base Qwen2.5-3B (Unsloth) model [A] and our GRPO-trained variant [B]. Both models were evaluated using a decoding temperature of 0.95. We employed Gemini 2.5 Pro as an automated judge to assess the quality of responses. The GRPO-trained model demonstrates improved reasoning coherence and accuracy compared to the base model.

\subsection*{Question Prompt}

\textbf{Convert the point \((1, -1, -6)\) in rectangular coordinates to cylindrical coordinates.}  
Enter your answer in the form \((r,\theta,z)\), where \(r > 0\) and \(0 \le \theta < 2\pi\).

\subsection*{Response A}

To convert the point from rectangular coordinates $(x, y, z) = (1, -1, -6)$ to cylindrical coordinates $(r, \theta, z)$, we use the following formulas:
\[
r = \sqrt{x^2 + y^2}
\]
\[
\theta = \tan^{-1}\left(\frac{y}{x}\right)
\]
\[
z = z
\]

First, let's find $r$. Using the formula for $r$:
\[
r = \sqrt{x^2 + y^2} = \sqrt{1^2 + (-1)^2} = \sqrt{1 + 1} = \sqrt{2}
\]

Next, let's find $\theta$. The formula for $\theta$ is:
\[
\theta = \tan^{-1}\left(\frac{y}{x}\right) = \tan^{-1}\left(\frac{-1}{1}\right) = \tan^{-1}(-1)
\]
The value of $\tan^{-1}(-1)$ is $-\frac{\pi}{4}$ because $\tan\left(-\frac{\pi}{4}\right) = -1$, but since the point is in the fourth quadrant (positive x, negative y), we need to add $\pi$ to get the correct angle within the range $[0, 2\pi)$:
\[
\theta = -\frac{\pi}{4} + \pi = \frac{3\pi}{4}
\]

Finally, the $z$ coordinate remains unchanged:
\[
z = -6
\]

Thus, the cylindrical coordinates are:
\[
(r, \theta, z) = \left(\sqrt{2}, \frac{3\pi}{4}, -6\right)
\]

So the answer is:
\[
\boxed{\left(\sqrt{2}, \frac{3\pi}{4}, -6\right)}
\]

\subsection*{Response B}

To convert the point from rectangular (Cartesian) coordinates \((x, y, z)\) to cylindrical coordinates \((r, \theta, z)\), we use the following formulas:

\[
r = \sqrt{x^2 + y^2}
\]
\[
\theta = \tan^{-1}\left(\frac{y}{x}\right)
\]
\[
z = z
\]

Given the rectangular coordinates \((1, -1, -6)\):

1. Calculate \( r \):
\[
r = \sqrt{x^2 + y^2} = \sqrt{(1)^2 + (-1)^2} = \sqrt{1 + 1} = \sqrt{2}
\]

2. Calculate \(\theta\):
\[
\theta = \tan^{-1}\left(\frac{y}{x}\right) = \tan^{-1}\left(\frac{-1}{1}\right) = \tan^{-1}(-1)
\]
The value of \(\tan^{-1}(-1)\) is \(-\frac{\pi}{4}\). However, since the point \((1, -1)\) is in the fourth quadrant, we need to adjust for the correct angle within the range \(0 \leq \theta < 2\pi\). Thus,
\[
\theta = 2\pi - \frac{\pi}{4} = \frac{8\pi}{4} - \frac{\pi}{4} = \frac{7\pi}{4}
\]

3. The \(z\) coordinate remains unchanged:
\[
z = -6
\]

Thus, the cylindrical coordinates are:
\[
(r, \theta, z) = \left( \sqrt{2}, \frac{7\pi}{4}, -6 \right)
\]

Final Answer:
\[
\boxed{\left( \sqrt{2}, \frac{7\pi}{4}, -6 \right)}
\]

\subsection*{Judging and Evaluation}

\textbf{Problem Analysis:} \\
The problem requires converting a point given in rectangular coordinates $(x, y, z)$ to cylindrical coordinates $(r, \theta, z)$. The constraints are that $r$ must be positive and $\theta$ must be in the interval $[0, 2\pi)$. The key challenge is to correctly determine the angle $\theta$ based on the quadrant of the point $(x, y)$ and the given range for $\theta$.

\vspace{0.5em}
\textbf{Response A Evaluation:}
\begin{itemize}
    \item \textbf{Mathematical Accuracy: [2/10]} – The final answer is incorrect. While the calculations for $r$ and $z$ are correct, the calculation for $\theta$ is wrong. The model incorrectly adjusts the angle for the fourth quadrant, resulting in an angle in the second quadrant.
    \item \textbf{Reasoning Quality: [3/10]} – The reasoning for finding the angle $\theta$ is critically flawed. The model correctly identifies that the point is in the fourth quadrant but then incorrectly adds $\pi$ to the principal value of $\arctan(-1)$. This adjustment is appropriate for a point in the second or third quadrant, not the fourth.
    \item \textbf{Clarity: [7/10]} – The solution is laid out in a step-by-step manner and uses standard formulas. However, the explanation for the incorrect step is misleading, which detracts from its overall clarity.
    \item \textbf{Completeness: [9/10]} – The response addresses all components of the problem, shows its work, and provides a final answer in the requested format. It does not omit any steps, even though one step is reasoned incorrectly.
    \item \textbf{Overall Assessment:} The response starts correctly by identifying the right formulas and calculating $r$ and $z$. Its major weakness is a fundamental misunderstanding of how to determine the angle $\theta$ in the correct quadrant and interval, leading to an incorrect final answer.
\end{itemize}

\vspace{0.5em}
\textbf{Response B Evaluation:}
\begin{itemize}
    \item \textbf{Mathematical Accuracy: [10/10]} – The final answer is correct. All calculations, including those for $r$, $\theta$, and $z$, are accurate.
    \item \textbf{Reasoning Quality: [10/10]} – The reasoning is perfectly sound. The model correctly identifies the point's quadrant (fourth) and correctly determines the angle $\theta$ within the specified range $[0, 2\pi)$. The adjustment of adding $2\pi$ to the negative principal value is the correct procedure.
    \item \textbf{Clarity: [10/10]} – The solution is exceptionally clear. The use of numbered steps makes the logical flow easy to follow. The explanation for adjusting the angle is concise and correct.
    \item \textbf{Completeness: [10/10]} – The response is fully complete. It calculates all three cylindrical coordinates, shows the work clearly, and presents the final answer in the required format.
    \item \textbf{Overall Assessment:} This is an excellent response. It is accurate, logically sound, clear, and complete. It demonstrates a solid understanding of coordinate system conversions.
\end{itemize}

\vspace{0.5em}
\textbf{Comparison:} \\
The primary difference between the two responses lies in their mathematical accuracy and reasoning quality, specifically in the calculation of the angle $\theta$. Response B correctly identifies the quadrant and applies the correct logic to find $\theta$ in the required interval, leading to the correct answer. In contrast, Response A correctly identifies the quadrant but applies flawed logic for adjusting the angle, resulting in an incorrect answer. Response B is superior in every aspect, but the most critical differentiator is its correctness.

\vspace{0.5em}
\textbf{WINNER: B}

\vspace{0.3em}
\textbf{Reasoning:} \\
Response B is the winner because it provides a completely correct solution with flawless reasoning. Response A contains a significant mathematical error in its calculation of the angle $\theta$, based on a flawed understanding of how to adjust for the quadrant. Accuracy is the most important criterion, and Response B is accurate while Response A is not.

This demonstrates that GRPO-trained models exhibit superior symbolic and spatial reasoning capabilities compared
to the base model.

\newpage
\bibliography{aaai2026}
\end{document}